%% file: main.tex
\documentclass[10pt]{article} %
\usepackage[accepted]{tmlr}

\input{math_commands.tex}

\usepackage{hyperref}       %
\usepackage{url}            %
\usepackage{booktabs}       %
\usepackage{amsfonts}       %
\usepackage{nicefrac}       %
\usepackage{microtype}      %
\usepackage{xcolor}         %
\usepackage{gensymb}
\usepackage{graphicx}
\usepackage{paralist}
\usepackage{comment}
\usepackage{wrapfig}
\usepackage{multicol}
\usepackage{multirow}
\usepackage{subcaption}
\usepackage{makecell}

\title{Learning Two-Step Hybrid Policy for Graph-Based Interpretable Reinforcement Learning}

\author{\name Tongzhou Mu \thanks{ Work done during an internship at Amazon Alexa AI.} \email t3mu@eng.ucsd.edu \\
      \addr Department of Computer Science and Engineering\\
      University of California San Diego
      \AND
      \name Kaixiang Lin \email kaixianl@amazon.com \\
      \addr Amazon
      \AND
      \name Feiyang Niu \email nfeiyan@amazon.com \\
      \addr Amazon
      \AND
      \name Govind Thattai \email thattg@amazon.com \\
      \addr Amazon
}

\newcommand{\update}[1]{\textcolor{black}{#1}}

\def\month{10}  %
\def\year{2022} %
\def\openreview{\url{https://openreview.net/forum?id=Ox5tmhFBrc}} %

\begin{document}

\maketitle

\begin{abstract}
We present a two-step hybrid reinforcement learning (RL) policy that is designed to generate interpretable and robust hierarchical policies on the RL problem with graph-based input. Unlike prior deep reinforcement learning policies parameterized by an end-to-end black-box graph neural network, our approach disentangles the decision-making process into two steps. The first step is a simplified classification problem that maps the graph input to an action group where all actions share a similar semantic meaning. The second step implements a sophisticated rule-miner that conducts explicit one-hop reasoning over the graph and identifies decisive edges in the graph input without the necessity of heavy domain knowledge. This two-step hybrid policy presents human-friendly interpretations and achieves better performance in terms of generalization and robustness. Extensive experimental studies on four levels of complex text-based games have demonstrated the superiority of the proposed method compared to the state-of-the-art. 
\end{abstract}

\input{sections/intro}
\input{sections/related_works}

\input{sections/method}

\input{sections/experiments}
\input{sections/conclusion}

\newpage

\bibliography{refs}
\bibliographystyle{tmlr}

\end{document}

%% file: math_commands.tex
\usepackage{amsmath,amsfonts,bm}

\def\eqref#1{equation~\ref{#1}}

\def\1{\bm{1}}

\DeclareMathAlphabet{\mathsfit}{\encodingdefault}{\sfdefault}{m}{sl}
\SetMathAlphabet{\mathsfit}{bold}{\encodingdefault}{\sfdefault}{bx}{n}

\DeclareMathOperator*{\argmax}{arg\,max}
\DeclareMathOperator*{\argmin}{arg\,min}

%% file: sections/intro.tex
\section{Introduction}

Recent years have witnessed the rapid developments of deep 
reinforcement learning across various domains such 
as mastering board games~\cite{silver2016mastering,schrittwieser2020mastering}
playing video games~\cite{mnih2015human}, and chip design~\cite{mirhoseini2021graph}, etc. The larger and complicated architecture of deep reinforcement 
learning models empowered the capabilities of resolving challenging tasks
while brought in significant challenges of interpreting the decision
making process of those complex policies~\cite{puiutta2020explainable}. This trade-off between 
performance and interpretability becomes an inevitable issue 
when DRL is applied to high stakes applications such as health care~\cite{rudin2019stop}. 
In this work, we focus on graph-based interpretable
reinforcement learning~\cite{zambaldi2018deep,waradpande2020graph} as the
graph representation is expressive in various domains including drug discovery~\cite{patel2020machine}, visual question answering~\cite{hildebrandt2020scene}, and embodied AI~\cite{chaplot2020neural,huang2019neural}, etc. \update{Especially in robotics and embodied AI, the scene graph~\cite{zhou2019scenegraphnet} of the environment is a general representation, which is widely used not only in building the environment ~\cite{puig2020watch, kolve2017ai2}, but also very effective when building the planning module~\cite{jia2022learning,min2021film}. } Another benefit of studying graph-based RL is that the graph structure can provide natural
explanations of the decision-making process without the necessity of introducing new \update{programs}~\cite{verma2018programmatically} or heavy domain knowledge~\cite{bastani2018verifiable} for interpretation.
Prior works in interpretable reinforcement learning~\cite{verma2018programmatically,madumal2020explainable,liu2018toward} either works on restricted
policy class~\cite{liu2018toward} that leads to downgraded performance, or the 
interpretablility~\cite{shu2017hierarchical,zambaldi2018deep} is limited. 
Another common issue of interpretable RL is that the provided explanation
is generally difficult to comprehend for non-experts~\cite{du2019techniques}. 

\update{
In this work, we seek to obtain a human-understandable interpretation for the decision making process in graph-based tasks. To better illustrate the desired interpretability in graph-based tasks, we borrow the ideas from the visual recognition community, where people interpret how CNN works by analyzing which region gets the most attention, like \cite{cnn_inter1,cnn_inter2}. Similarly, in graph-based tasks, we want to \textbf{figure out the most crucial sub-graphs (or edges) used in decision making}. 
\footnote{\color{black}The idea of the crucial sub-graph is also similar to the crucial point set used in PointNet\cite{qi2017pointnet}, which is essentially a graph neural network.} 
We refer to this kind of interpretation as “interpretability” in graph-based RL, not only because this kind of “interpretability” is similar to the one in the visual recognition community, but also because it is easily understood by humans.
A crucial sub-graph also has several desirable characteristics of explanation defined by \cite{molnar2020interpretable}, e.g., it is contrastive and selected.
}

\update{To obtain the desired interpreability and} resolve the challenges mentioned above, we propose a novel two-step hybrid decision-making process for general
deep reinforcement learning methods with graph input. 
Our approach is inspired by the observation of human decision-making. When
confront complicated tasks that involve expertise from multiple
domains, we typically identify which domain of expert we would like
to consult first and then search for specific knowledge or solutions to solve the 
problem. Recognizing the scope of the problem significantly
reduces the search space of downstream tasks, which leads to a more simplified
problem compared to find the exact solution in all domains directly.  
As an analogy of this procedure, we disentangle a complicated deep
reinforcement learning policy into a classification 
the problem for problem type selection and rule-miner.
The classification establishes a mapping from complex graph input 
into an action type, which handles high-order logical interactions among 
node and edge representations with graph neural network. 
The rule miner conducts explicit one-hop reasoning over the graph
and provides user-friendly selective explanations~\cite{du2019techniques} by
mining several decisive edges. 
This two-step decision making is essential not only for providing 
interpretability, but also for generalization and robustness. 
It is intuitive to see that the simplified classification is easier
to achieve better generalization and robustness 
than the original complicated RL policy. 
Furthermore, the rule-miner identifying key edges in the graph is
much more robust to the noisy perturbations on the irrelevant 
graph components.

\update{
Our contributions can be summarized as below:
\begin{itemize}
    \item We propose a general framework for two-step hybrid decision making, which is able to combine neural network and rule-based method together to make a decision while yielding human-friendly interpretations.
    \item We instantiate our proposed framework in the setting of text-based games with graph input by proposing a novel rule miner for knowledge graphs. 
    \item Extensive experiments on several text-based games~\cite{cote2018textworld} demonstrated that the proposed approach achieves a new state-of-the-art performance for both generalization and robustness.
\end{itemize}
}

%% file: sections/related_works.tex
\section{Related Works}

\subsection{Interpretable RL}
Instead of using deep RL as a black box, researchers also worked on making deep RL more interpretable.  \cite{mott2019towards, zambaldi2018deep} introduce some sorts of attention \update{mechanism} into policy networks and explain decision making process by analyzing attention weights. \cite{verma2018programmatically} combines program synthesis and PIDs in classic control to solve continuous control problems, while keep the decision making interpretable by explicitly writing the program. Combining symbolic planning and deep RL has similar effects, like \cite{lyu2019sdrl}. Tree-based policies \cite{bastani2018verifiable} are favorable in interpretable RL, since they are more human-readable and easy to verify. \update{\cite{bonaert2020explainable} introduces goal-based interpretability, where the agent produces goals which show the reason for its current actions.} However, many existing works of interpretable RL sacrifices the performance for the interpretability. In contrast, our method achieves even better generalization and robustness while providing interpretability.

\subsection{Hierarchical RL}
Hierarchical RL focuses on decomposing a long-horizon tasks into several sub-tasks, and applying a policy with hierarchical structure to solve the task. \cite{vezhnevets2017feudal, frans2018meta} try to solve generic long-horizon tasks with a two-level policy, where a high-level policy will generate a latent sub-goal for low-level policies, or directly select a low-level policy, and the low-level policy is responsible for generating the final action. In some robot locomotion tasks, people usually manually define the choices of sub-goals to boost the performance and lower the learning difficulty, like \cite{nachum2018data, levy2018learning}. \update{In Multi-agent RL, hierarchical structure is a commonly adopted policy structure to provide certain interpertability~\cite{gupta2021uneven,mahajan2019maven,wang2020rode}. However, those approaches can not explain how specific decision is associated with the state space as their low-level policy is still a black box policy.}
Our two-step hybrid policy is reminiscent of the two-level policy architecture in hierarchical RL. Unlike the high-level policy in hierarchical RL is mainly used to generate a sub-goal, the action pruner in our hybrid policy is also used to reduce the number of action candidates. And the motivations of our method and hierarchical RL are also different: hierarchical RL targets at decomposing a long-horizon tasks into several pieces, but our method focuses on interpertability, generalizability and robustness.

\subsection{Refactoring Policy}
Given a trained neural network-based policies, one may want to refactor the policy into another architecture, so as to improve the generalizibility or interpretablity. And this refactoring process is usually done by imitation learning. The desired architectures of the new policy can be decision tree \cite{bastani2018verifiable}, symbolic policy \cite{landajuela2021discovering}, a mixture of program and neural networks \cite{sun2018neural}, a graph neural networks \cite{mu2020refactoring}, \update{or a hierarchical policy \cite{bonaert2020explainable}}. In our method, we also use similar techniques to refactor a reinforcement learning policy into a two-step hybrid policy.

\subsection{Combine Neural Networks with Rule-Based Models}
Neural networks and rule-based models excel at different aspects, and many works have explored how to bring them together, like \cite{chiu1997combining, goodman1990rule, ray2020mixed, okajima2019deep, greenspan1992combined}. There are various ways to combine neural networks with rule-based models. For example, \cite{wang2019gaining} combines them in a "horizontal" way, i.e., some of the data are assigned to rule-based models while the others are handled by neural networks. In contrast, our work combines neural networks and rule-based models in a "vertical" way, i.e., the two models collaborate to make decisions in a two-step manner.

%% file: sections/method.tex
\section{A General Framework for Two-Step Hybrid Decision Making}

In this section, we first describe our problem setting including key assumptions and a general framework that formulates the decision-making process in a two-step manner. 

We consider a discrete time Markov Decision Process (MDP) with a discrete
state space $\mathcal{S}$ and the action space $\mathcal{A}$ is finite. 
The environment dynamics is denoted as $\mathbf{P} = \{p(s'|s, a), \forall s, s' \in \mathcal{S}, a \in \mathcal{A}\}$. 
Given an action set $\mathcal{A}$ and the current state $s_t \in \mathcal{S}$, our goal is to learn a policy $(\pi)$ that select an action $a_t\in \mathcal{A}$ to maximize the long-term reward $\mathbf{E}_{\pi}[\sum_{i=1}^T r(s_i, a_i)]$.  Assume we are able to group the actions into several mutual exclusive action types ($A_k$) according to its semantic meanings. 
More concretely,  the $k$-th action type $A_k = \{a_k^1,...,a_k^n\}$ denotes a subset of actions in original action space $A_k\subseteq A$.  
Then we have $A_1,A_2,..,A_K \subseteq \mathcal{A}, A_i\cap A_j=\emptyset (i\neq j), \cup_{i=1}^K A_i = \mathcal{A}$. It is worth noting that
the number of action type $K$ is usually much smaller than original actions ($K \ll |\mathcal{A}|$).

Let the policy $\pi=\langle f_p, f_s \rangle$ represented by a hybrid model that consists of action pruner $f_p$ and action selector $f_s$. The action pruner is used to prune all available action candidates to a single action type, i.e., $f_p(s_i)=k$, where $k$ is the index of chosen action type and $A_k \in \{A_1, A_2, .., A_K\}$. Then the action selector is used to select a specific action given the action type chosen by the action pruner, i.e., $f_s(s_i, A_k)=a_i$, where $a_i\in A_k$ and $k = f_p(s_i)$.

Intuitively, this design intends to disentangle different phases in decision-making process to two different modules. On the one hand, determining the action type typically involves high-order relationships and needs a model with strong expressive power, then neural network is a good candidate in this regard. On the other hand, selecting an action within a specific action type can resort to rule-based approaches, which is essential in providing strong intrepretability, generalizability and robustness.

\begin{figure}[htbp]
    \centering
    \includegraphics[width=\linewidth]{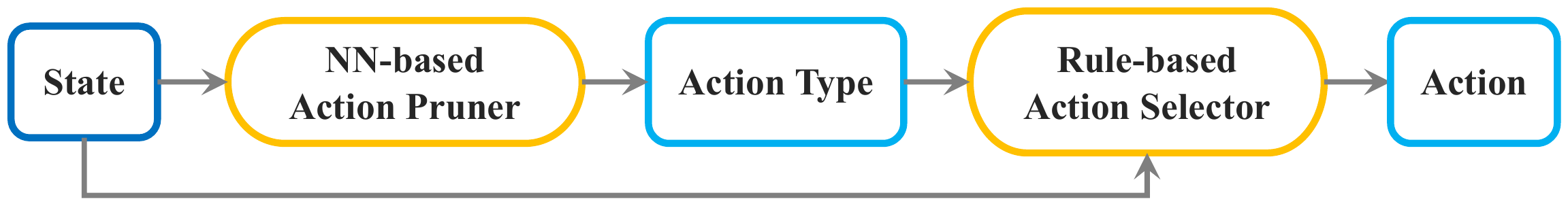}
    \caption{Overall pipeline of the two-step hybrid decision making.}
    \label{fig:overall_pipeline}
\end{figure}

Figure~\ref{fig:overall_pipeline} shows the overall pipeline of our two-step hybrid decision making. The agent will receive a state $s_i$ at each time step $i$.
At time step $i$, we will first call the action pruner $f_p(s_i)$ to select the action type $A_{k_i}$. Then the rule-based action selector $f_s(s_i, A_{k_i})$ will take as inputs the current state $s_i$ and the action type $A_{k_i}$ given by action pruner to select the specific action to be executed in this step.

\section{Two-Step Hybrid Policy for Text-Based Games with Graph Input}

\subsection{Overview}
In this section, \update{we first discuss our considerations for investigating the interpretability of graph-based tasks and the reasons for using text-based games as our testbed. Firstly, graph is a powerful data structure, and it has been widely adopted in many real-world applications such as drug discovery~\cite{patel2020machine}, visual question answering~\cite{hildebrandt2020scene}, and embodied AI~\cite{chaplot2020neural,huang2019neural}, etc. Meanwhile, the discrete nature and the explicit semantic meaning of graphs (e.g., the relationship of related nodes) provide an easily understandable format for elaborating the decision-making process even for a non-expert. However, the interpretability of graph-based RL tasks is still rarely explored. On the other hand, it is challenging to provide interpretability without any assumption on the action space, even for graph-based RL. We focus on tasks with hierarchical/compositional action spaces as a starting point, which is a common setting in embodied AI. Considering the text-based scene graph has been a general representation of embodied AI tasks~\cite{shridhar2020alfworld},} we instantiate our proposed framework in the setting of text-based games~\cite{cote2018textworld} with graph input. 
In text-based games, the agent receives a knowledge graph (as shown in Figure~\ref{fig:kg_example}) that describes the current state of the environment and the task to be completed, e.g., the task can be represented by several edges like (``potato", ``needs", ``diced"). Our goal is to learn a policy that maps the input knowledge graph to an action from the provided action set $\mathcal{A}$. Each action $a_j\in\mathcal{A}$ is a short sentence, e.g., ``take apple from fridge". 

In this setting, we use graph neural networks (GNNs) as the action pruner $f_p$, and a rule-based model as the action selector $f_s$. We will elaborate the details of the action pruner and the action selector in Sec~\ref{sec:gnn_action_pruner} and Sec~\ref{sec:rule_model}, respectively. \update{A brief description of GNNs can be found in Sec \ref{sec:gnn_bg}.}

Training the GNN-based action pruner and the rule-based action selector by reinforcement learning is nontrivial since the whole pipeline is not end-to-end differentiable. Therefore, we propose to learn both models separately from a demonstration dataset, and the demonstration dataset can be obtained by a trained reinforcement learning agent. We will elaborate the details in  Sec~\ref{sec:demonstration-acquisition}. This process is inspired by existing works like \cite{bastani2018verifiable,landajuela2021discovering,sun2018neural, mu2020refactoring}, where policies trained by reinforcement learning can be refactorized into other forms.

\update{
\subsection{Backgrounds on Graph Neural Networks}
\label{sec:gnn_bg}
We briefly describe a commonly used class of GNNs that encompasses many state-of-art networks, including GCN \cite{gcn}, GAT \cite{gat}, GIN \cite{gin},  etc. The class of GNNs generalizes the message-passing framework \cite{message} to handle edge attributes. Each layer of it can be written as 
$$
h^{l+1}_v = f_\theta^l(h_v^l, \{x_e\colon e\in\mathcal{N}_E(v)\}, \{h_{v'}^l\colon v'\in\mathcal{N}_V(v)\})
$$,
where $h_v^l$ is the node feature vector of node $v$ at layer $l$.  $\mathcal{N}_V(v)$ and $\mathcal{N}_E(v)$ denote the sets of nodes and edges that are directly connected to node $v$ (i.e. its 1-hop neighborhood). 
$f_\theta^l$ is a parameterized function, which is usually composed of several MLP modules and aggregation functions (e.g. sum/max) in practice. 
}

\subsection{Preparing Datasets for Learning Two-Step Hybrid Policy}
\subsubsection{Demonstration Acquisition}
\label{sec:demonstration-acquisition}
Given a set of interactive training environments, the goal of this step is to obtain a demonstration dataset, which contains state-action pairs from a policy achieving high reward in the training environment.
Specifically, we aim at generating a demonstration dataset $\mathcal{D}=\{(s_i, \pi(s_i))\}_{i=1}^N$, where $s_i$ is the input state (a knowledge graph in our cases) from the training environments, and $\pi(s_i)$ is the output of demonstration policy on the state $s_i$. The representation of $\pi(s_i)$ is flexible: it can be an action, the logits, or any latent representation, which indicates the demonstration action distribution. This dataset will be used for learning our two-step hybrid policy.

We refer the policy used to generate demonstration as \textit{teacher policy}. It can be obtained in any way, as long as it provides reasonable and good supervision on how to solve the task in the training environments, and it is not necessary to have strong interpretablity, generaliziabilty or robustness. Therefore, deep reinforcement learning is a great tool to learn such a policy.

After obtaining the teacher policy, we use it to interact with the training environments to collect states and label them with the output of the teacher policy. During the interaction, we add action noise to perturb the state distribution in the demonstration dataset. A more diversified state distribution is beneficial to the performance of our two-step hybrid approach. 

\subsubsection{Action Grouping and Demonstration Splits}
\label{sec:action_group}
As mentioned above, the action pruner is responsible for selecting the correct type of actions for the current state, so that the action candidates will be pruned to a smaller set. Then, a rule-based model will work on this specific action type to select the action to be executed. Therefore, we \update{need} to label each state with the action type. Since the action for each state is already in the dataset, we just need to convert each action into an action type. In other words, we need to group all possible actions into several action types.  

In text-based games, each action is a short sentence with semantic meaning, e.g., ``take apple from fridge". Intuitively, we can group the actions by their semantic meaning. For example, if the actions are ``take apple from fridge", ``slice potato with knife", ``dice cheese with knife", ``cook carrot with oven", and ``cook onion with stove", we can group them into three types: ``take" actions, ``cut" actions, and ``cook" actions. And we make sure all the actions with the same action type follow the same template, e.g., the template of ``take" action can be written as ``take \underline{object} from \underline{receptacle}", where \underline{object} and \underline{receptacle} can be instantiated by appropriate words.

Grouping actions by semantic meanings also aligns with our motivation:
1) After we obtained $K$ group of actions, we substantially alleviate the workload of neural networks from learning a policy that maps states to an action in a varying size and high-dimensional action space to a simplified classification with a much smaller number of classification categories. This simplification could greatly reduces the sample-complexity.
2) Furthermore, \update{this step also creates a much smaller search space to enable the rule-miner to search within a feasible candidates set. }

Formally, given a dataset $\mathcal{D}=\{(s_i, \pi(s_i))\}_{i=1}^N$, we will generate an action type for each state, policy output ($s_i,\pi(s_i)$) pair. The action type $k_i$ is determined by $k_i = h(\pi(s_i))$, where $h(\pi(s_i))$ is a domain-specific function which takes the semantic meaning of the action into consideration. As a result of the clustering process, we will get a new demonstration dataset with action types $\mathcal{D}=\{(s_i, \pi(s_i), k_i)\}_{i=1}^N$. In addition, we can also split the demonstration dataset based on the action types to get $K$ subset of the original demonstration dataset, $\{\mathcal{D}_1, \mathcal{D}_2, ..., \mathcal{D}_K\}$, where $\mathcal{D}_k=\{s_i,\pi(s_i) | h(\pi(s_i))=k \}$.

\subsection{GNN-based Action Pruner}
\label{sec:gnn_action_pruner}
The action pruner needs to output an action type based on the input knowledge graph, so it is essentially a classifier.
Given the demonstration dataset $\mathcal{D}=\{(s_i, k_i)\}_{i=1}^N$ obtained in the last step, we want to train a classifier $f_p(s;\theta)=k$, where $k\in\{1,2,...,K\}$ is an action type. This is a conventional classification problem which can be solved by minimizing cross entropy loss:
$$\theta = \argmin_\theta -\sum_i\sum_{j=1}^K k_i^j \log(f_\theta^j(s_i)), $$
where $f_\theta(s_i)$ outputs a probability distribution over the $K$ action types, $f_\theta^j(s_i)$ denotes the $j$-th action type's probability. $k_i^j \in \{0, 1\}$ denotes denotes whether the action type $j$ was chosen in the demonstration dataset at state $i$. 

\subsection{Rule-Based Action Selector}
\label{sec:rule_model}

\subsubsection{Abstract Supporting Edge Sets}

When the input is a knowledge graph, the action is naturally strongly correlated with some critical edges. For example, (``potato", ``needs", ``diced") and (``potato", ``in", ``player") can lead to the action ``dice potato".
We refer those decisive edges correlating to an action as the \textit{supporting edge set} of this action. 
Since we have grouped actions by their semantic meanings, actions within each action type are actually supported by similar edges. For example, ``dice potato" is supported by (``potato", ``in", ``player"), and ``dice apple" is supported by (``apple", ``in", ``player"). As mentioned in Sec~\ref{sec:action_group}, each action type comes with an action template like ``dice \underline{object}". Based on the action template, we can perform some sorts of abstraction. For example, given an input knowledge graph labeled with action ``dice apple", we can replace all the ``apple" appearing in the graph edges and the action with an abstract name ``\underline{object}". Then we can say the action ``dice \underline{object}" is essentially supported by (``\underline{object}", ``in", ``player"), where the two ``\underline{object}" should be instantiated by the same word. Under this kind of abstraction, different actions within the same action type can share a same \textit{abstract supporting edge set} which contains edges with abstract names.

The \textit{abstract supporting edge set} indicates the decisive edges for an action type, and it can be instantiated for each specific action. For example, to check whether the action ``dice apple" should be executed, the abstract edge (``\underline{object}", ``in", ``player") will be instantiated to edge (``apple", ``in", ``player"). Then, the existence of (``apple", ``in", ``player") in input knowledge graph becomes an evidence for selecting the action ``dice apple". We aim at finding an abstract supporting edge set for each action type, and it will be used during inference.

\subsubsection{Mine Abstract Supporting Edge Sets from Demonstrations}
\label{sec:mine_rules}

Finding the abstract supporting edge set for each action type is actually a rule mining process. There are several off-the-shelf rule miners like FP-Growth \cite{fp_growth}, Apriori \cite{agrawal1994fast}, Eclat \cite{zaki2000scalable}, etc. , but they are not designed for knowledge graphs. Thus, we propose a simple yet effective rule miner for our setting to discover the supporting edge sets.

To find the \textbf{A}bstract \textbf{S}upporting \textbf{E}dge set $\text{ASE}(A_k)$ for each action type $A_k$, we designed a numerical statistic that is intended to reflect the importance of an edge when taking an action $a\in A$, and this numerical statistic is inspired by tf-idf \cite{rajaraman2011mining}. 
Formally, for action type $A_k$, \update{we have a corresponding subset of demonstration dataset  $\mathcal{D}_k=\{s_i\}$ (ignoring $\pi_i$ here) where all the actions are within the action type $A_k$}. Under the abstraction mentioned above, we can count the edge frequency for every (abstract) edge $e$ in $\mathcal{D}_k$: $$freq_k(e)=\frac{|\{s_i| s_i\in \mathcal{D}_k, e\in s_i \}|}{|\mathcal{D}_k|}_.$$
Similarly, we can also count the edge frequency for the entire demonstration dataset:
$$freq(e)=\frac{|\{s_i| s_i\in \mathcal{D}, e\in s_i \}|}{|\mathcal{D}|}_.$$
And we can define an importance score of an edge w.r.t to the action type $A_k$:
$$I_k(e)=freq_k(e)\cdot \log(\frac{1}{freq(e)}),$$
where the term $freq_k(e)$ is similar to the term-frequency (tf), and the term $log(\frac{1}{freq(e)})$ is similar to the inverse document frequency (idf).

Then we can get the $\text{ASE}(A_k)$ by selecting the edges with the importance higher than a threshold, i.e., $\text{ASE}(A_k)=\{e|I_a(e)>\tau\}$, where $\tau$ is a hyperparameter shared across all action types.

\begin{figure}[tbp]
  \centering
      \includegraphics[width=\linewidth]{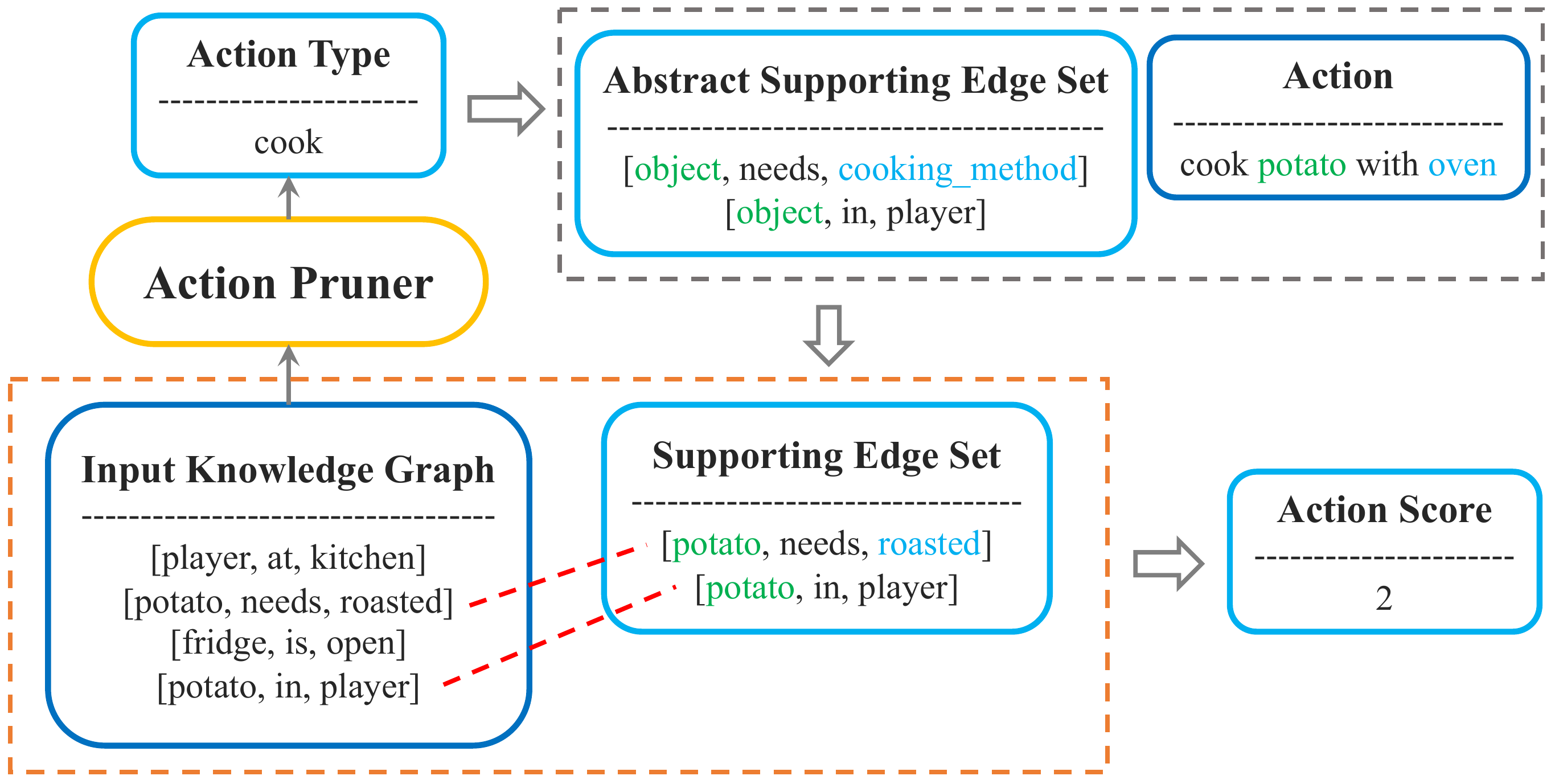}
    \caption{An example of using supporting edge set to score an action. 
    Based on the knowledge graph, the action pruner first predict the action type (e.g., cook) that needs to be taken at current state, then instantiate the abstract supporting edge set to a concrete supporting edge set. Comparing the input knowledge graph with the supporting edge set, we can compute action score for each action and select the action with highest score accordingly. 
    }
\label{fig:inference_example}
\end{figure}

\subsubsection{Inference Based on Supporting Edges}

During inference, we use the supporting edge sets to score each action within the action type provided by action pruner, and select the action with the highest score. Figure~\ref{fig:inference_example} shows a concrete example of using supporting edge set to score an action.

Firstly, the action pruner outputs an action type based on the input knowledge graph, and we can retrieve the abstract supporting edge set of this action type.
Secondly, given an action within the action type, we can instantiate the abstract supporting edge set to a specific supporting edge set by replacing the abstract names by the concrete words according to the action, e.g, (``\underline{object}", ``in", ``player") will be instantiated to (``potato", ``in", ``player") if the action is ``cook potato with oven".
Finally, we can compare the input knowledge graph with the supporting edge set of each action to figure out which supporting edge set is best covered by the input knowledge graph. The number of overlapped edges between the supporting edge set and the the input knowledge graph will be regarded as the score of the action.

Formally, the inference process of our rule-based action selector can be described as follows
$$
f_s(s, A) = \argmax_{a\in A} | s \cap \text{SE}(a)|,
$$
 where $\text{SE}(a)$ is the supporting edge set associated with action $a$, which is obtained by instantiating the abstract supporting edge set of action type $A$.

%% file: sections/experiments.tex
\section{Experiments}

\subsection{Dataset Setup}

\begin{table}[htbp!]
\centering
{\color{black}
\begin{tabular}{c|ccccccc}
\toprule
\begin{tabular}[c]{@{}c@{}}Difficulty\\ Level\end{tabular} & \begin{tabular}[c]{@{}c@{}}Recipe\\ Size\end{tabular} & \begin{tabular}[c]{@{}c@{}}Number of\\ Locations\end{tabular} & \begin{tabular}[c]{@{}c@{}}Need\\ Cut\end{tabular} & \begin{tabular}[c]{@{}c@{}}Need\\ Cook\end{tabular} & \begin{tabular}[c]{@{}c@{}}Number of\\ Action Candidates\end{tabular} & \begin{tabular}[c]{@{}c@{}}Number of\\ Objects\end{tabular} &
\begin{tabular}[c]{@{}c@{}}Number of\\ Sub-tasks\end{tabular} \\
\midrule
1                                                          & 1                                                     & 1                                                             & Yes                                                & No                                                  & 11.5                                                                  & 17.1         & 4                                               \\
2                                                          & 1                                                     & 1                                                             & Yes                                                & Yes                                                 & 11.8                                                                  & 17.5            &5                                            \\
3                                                          & 1                                                     & 9                                                             & No                                                 & No                                                  & 7.2                                                                   & 34.1               & 3                                         \\
4                                                          & 3                                                     & 6                                                             & Yes                                                & Yes                                                 & 28.4                                                                  & 33.4          & 11    \\
\bottomrule
\end{tabular}
}
\caption{TextWorld games statistics (averaged across all games within a difficulty level). The games and statistics are generated by \cite{gata}.}

\label{tab:textworld_stat}
\end{table}

We evaluate our method on TextWorld, which is a framework for designing text-based interactive games. More specifically, we use the TextWorld games generated by GATA~\cite{gata}. In these games, the agent is asked to cook a meal according to given recipes. It requires the agent to navigate among different rooms to locate and collect food ingredients specified in the recipe, process the food ingredients appropriately, and finally cook and eat a meal.

\begin{wrapfigure}{r}{6cm}
    \vspace{-2 cm}
    \centering
    \captionsetup{justification=centering}
    \includegraphics[width=5.5cm]{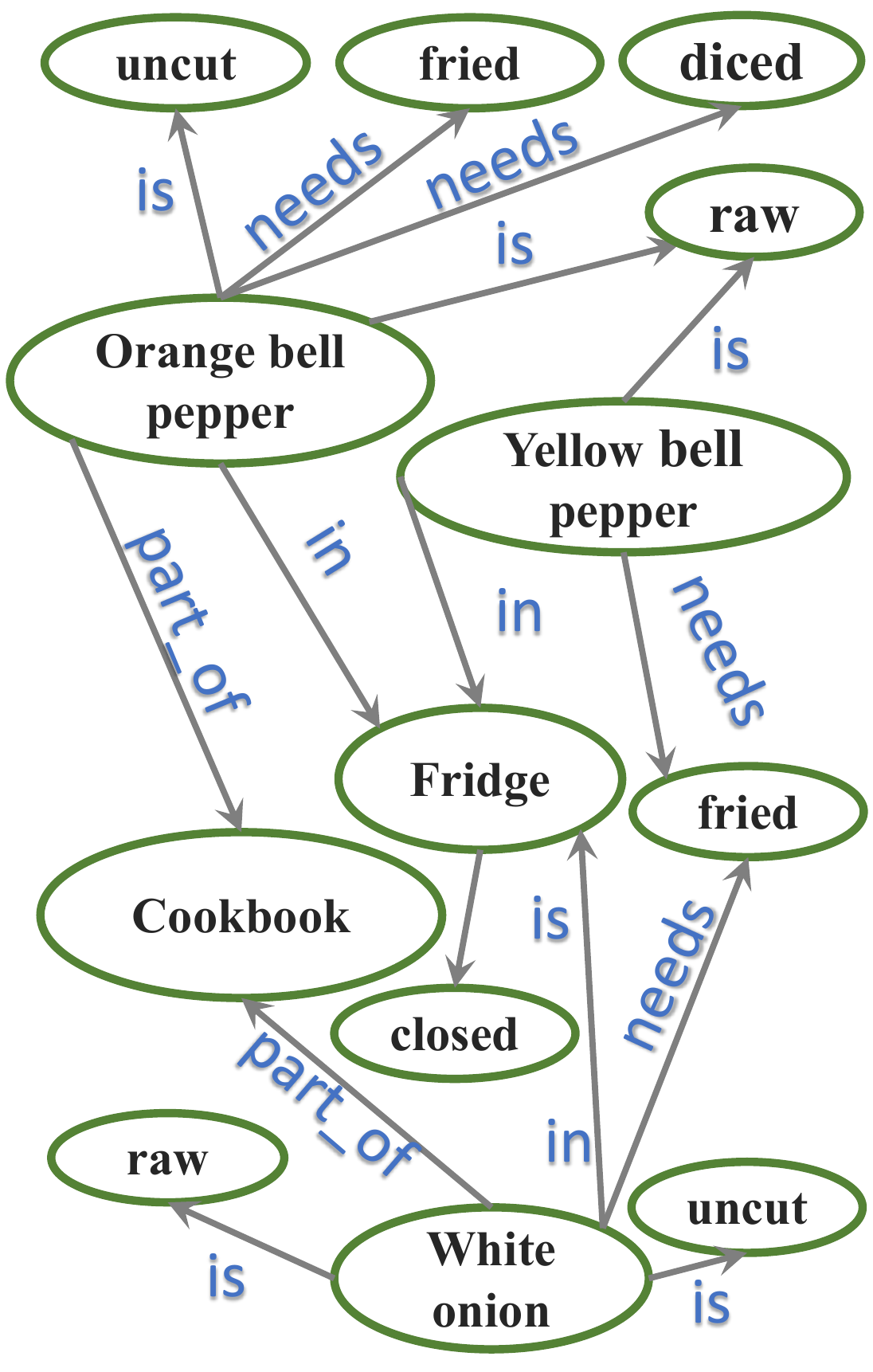}
    \caption{Visualization of a part of an input knowledge graph in TextWorld.}
    \label{fig:kg_example}
    \vspace{-0.2cm}
\end{wrapfigure}

The state received by the agent is a knowledge graph describing all the necessary information about the game. All the nodes and edges in the knowledge graph are represented in text. Figure~\ref{fig:kg_example} shows a partial example of input knowledge graph. The actions are also represented in text. Note that the number of available actions vary from state to state, so most of of existing network architecture used deep reinforcement learning cannot be directly used here.

The games have four different difficulty levels, and each difficulty level contains 20 training, 20 validation, and 20 test environments, which are sampled from a distribution based on the difficulty level. The higher the difficulty levels are, the more complicated recipe will be and the more rooms food ingredients will be distributed among. \update{For easier games, the recipe might only require a single ingredient, and the world is limited to a single location, whereas harder games might require an agent to navigate a map of six locations to collect and appropriately process up to three ingredients.} Statistics of the games are shown in Table~\ref{tab:textworld_stat}. 
For evaluating model generalizability, we select the top-performing agent on validation sets and report its test scores; all validation and test games are unseen in the training set. 

\subsection{Implementation Details}

\subsubsection{Process Knowledge Graphs by GNNs}
Graph neural networks (GNNs) serve as backbone networks in both our RL teacher policy and the action pruner in our two-step hybrid policy. More specifically, we use the Relational-GCN to take edge attributes into consideration\update{, and the Relational-GCN is also the backbone architecture used in GATA \cite{gata}, which is the baseline method in the TextWorld games we used.}
\update{
Actually, the choice of the network architecture is not very important in our method. This is because both the RL teacher policy and the action pruner are only used to fit the training environments or the demonstration dataset. We did not really expect them to have strong interpretability, because this is the job of the rule-based action selector. Therefore, most graph-based networks with sufficient capacity should be good in our cases. We will also perform an ablation study about network architecture in Sec. \ref{sec:ablation}.
}

The relations and nodes of the input knowledge graphs are initially represented in text (words or short sentences). We use fastText \cite{fasttext}, pre-trained on Common Crawl, to convert each token to an embedding, and then average all the token embeddings to obtain the mean fastText embedding of the entire text string. These mean fastText embeddings are pre-computed and fixed during training. For each relation and node, we append their corresponding mean fastText embedding with a trainable embedding vector. The concatenated embeddings are then used as the numerical representation of each relation and node. \update{Finally, a graph with numerical node/relation features is passed through GNNs to get the latent embedding of the entire input knowledge graph}.

\subsubsection{Demonstration Acquisition}

To collect demonstration dataset, we first train a teacher policy by DQN~\cite{mnih2015human} in the training environments, which can converge to a near-optimal solution. To adopt variable number of available actions, we let the policy network take the state and an available action as input and output a score for this action. And we can select the action with maximum score as the final output.

The trained teacher policy is used to collect 300K samples through the interaction with the environment, and label them with the taken actions, as illustrated in Sec~\ref{sec:demonstration-acquisition}. When collecting the demonstration dataset, we use $\epsilon$-greedy exploration strategy to increase the diversity of states.

\subsection{Results}

\subsubsection{Interpretability}

\begin{table}[!htbp]
\centering
\begin{tabular}{c|cccc|ccc}
\toprule
                                                                            & \multicolumn{4}{c|}{Action Type: Cut}                 & \multicolumn{3}{c}{Example Action}          \\
\midrule
Action                                                                      & \multicolumn{4}{c|}{\underline{verb} \underline{object} with knife}           & \multicolumn{3}{c}{slice potato with knife} \\
\midrule
\multirow{5}{*}{\begin{tabular}[c]{@{}c@{}} \\ \\Discovered\\ Supporting\\ Edges\end{tabular}} & Head   & Relation & Tail          & Importance & Head        & Relation      & Tail          \\
\midrule
                                                                            & \underline{object} & in       & player        & 1.7004           & potato      & in            & player        \\
                                                                            & \underline{object} & needs    & \underline{verb\_passive} & 1.6659           & potato      & needs         & sliced        \\
                                                                            & \underline{object} & part\_of & cookbook      & 1.0058           & potato      & part\_of      & cookbook      \\
                                                                            & \underline{object} & is       & uncut         & 0.9837           & potato      & is            & uncut    \\
\bottomrule
\end{tabular}
\caption{Discovered abstract supporting edges of the ``cut" action type in difficulty level 1 of TextWorld environments. The importance of each edge is computed by the metric mentioned in Sec~\ref{sec:mine_rules}. The right column gives an example action and the corresponding supporting edge set, which is obtained by instantiating the abstract supporting edge set. The underlined words are abstract names for some certain words. For example, ``verb" in action can be instantiated to ``slice", ``dice" or ``chop", and ``verb\_passive" can be instantiated to ``sliced", ``diced" or ``chopped" accordingly.}
\label{tab:textworld_inter}
\end{table}

The interpretablity of our two-step policy is two-fold: 1) the transparent two-step decision making process; 2) the rule-based models make decisions in a way which is easy to interpret by human. 

Table~\ref{tab:textworld_inter} shows some representative rules discovered by our rule miners. We observed that all of the four discovered abstract supporting edges are indeed prerequisites of the ``cut" actions. In particular, the abstract supporting edge (``object", ``needs", ``verb\_passive") is a crucial prerequisite for the agent to select the correct verb in the ``cut" actions. For example, if the input knowledge graph contains the edge (``potato", ``needs", ``sliced"), then the action ``slice potato with knife" will get one more score than others because this edge is in the supporting edge set of action ``slice potato with knife".

Since we clearly see the rule-based model makes decisions based on these rules, it is not hard to check whether this model works in a correct way. In this way, the agent can certainly select the correct ``cut" action even in unseen test environments.

\subsubsection{Generalization}

\begin{table}[!htbp]
\centering
\begin{tabular}{c|cccc|cccc}
\toprule
          & \multicolumn{4}{c|}{Training}            & \multicolumn{4}{c}{Test}    \\
\midrule
Difficulty & 1    & 2           & 3           & 4             & 1   & 2            & 3    & 4             \\
\midrule
GATA-GTF      & 98.6 & 58.4          & 95.6        & 36.1         & 83.8 & 53          & 33.3        & 23.6          \\ 
Vanilla RL   & 100 & 100 & 98.3 & 100 & 83.8 & 68 & 50 & 30.9 \\ 
Ours   & 100 & 100 & 100 & 65.5 & \textbf{100} & \textbf{100} & \textbf{51.7} & \textbf{49.7}\\
\bottomrule
\end{tabular}
\caption{Evaluation results on both training environments and test environment in TextWorld. The numbers show the agent's normalized scores.}
\label{tab:textworld_gen}
\end{table}

We compare with two baselines: GATA-GTF and vanilla RL. 
GATA-GTF is a variant from GATA \cite{gata}, and it uses the same ground-truth graph input as us. GATA-GTF processes the input graphs by relational graph convolutional networks (R-GCNs) \cite{rgcn}, and the whole pipeline is trained by DQN \cite{mnih2015human}.
Vanilla RL is essentially the same method with GATA-GTF, but implemented by ourselves. With better implementation and careful hyperparameter tuning, it performs better than the implementation released by the GATA authors. And vanilla RL also serves the teacher policy which used to generate demonstrations for our method.

Table \ref{tab:textworld_gen} shows the normalized scores of different methods on both training environments and test environment in TextWorld. The result of GATA-GTF is obtained from its paper \cite{gata}.
In all the environments, our agent achieves better generalization performance the vanilla RL (which is our RL teacher) and GATA-GTF baselines. Our agent can generalize pretty well to the unseen test environments in all difficulty levels, while the performance of GATA-GTF and vanilla RL performs poorly in unseen test environments.

\subsubsection{Robustness}

\begin{table}[]
\centering
\setlength\tabcolsep{4pt} %
\begin{tabular}{cc|cc|cc|cc|cc}
\toprule
\multicolumn{2}{c|}{Noise} & \multicolumn{8}{c}{Difficulty}      \\
\midrule
&      & \multicolumn{2}{c|}{1}                 & \multicolumn{2}{c|}{2}                  & \multicolumn{2}{c|}{3}          & \multicolumn{2}{c}{4}                   \\
\midrule
Add         & Drop       & RL                & Ours              & RL                & Ours               & RL        & Ours               & RL                 & Ours   \\

\midrule
0.2           & 0              & \textbf{100(0\%)} & \textbf{100(0\%)} & \textbf{100(0\%)} & \textbf{100(0\%)}  & 96(-1\%)  & \textbf{98(-1\%)}  & 38(-61\%)          & \textbf{64(-1\%)}  \\
0.2           & 0.03           & \textbf{100(0\%)} & 96(-3\%)          & 81(-19\%)         & \textbf{98(-1\%)}  & 80(-18\%) & \textbf{98(-1\%)}  & 41(-58\%)          & \textbf{51(-21\%)} \\
0.2           & 0.06           & \textbf{100(0\%)} & 92(-7\%)          & \textbf{93(-6\%)} & 82(-18\%)          & 86(-11\%) & \textbf{93(-6\%)}  & \textbf{44(-55\%)} & 41(-36\%)          \\
0.4           & 0              & \textbf{100(0\%)} & \textbf{100(0\%)} & 85(-15\%)         & \textbf{100(0\%)}  & 65(-33\%) & \textbf{91(-8\%)}  & 19(-80\%)          & \textbf{58(-11\%)} \\
0.4           & 0.03           & \textbf{100(0\%)} & \textbf{100(0\%)} & 77(-23\%)         & \textbf{90(-9\%)}  & 71(-27\%) & \textbf{91(-8\%)}  & 20(-79\%)          & \textbf{53(-18\%)} \\
0.4           & 0.06           & \textbf{100(0\%)} & 96(-3\%)          & 78(-22\%)         & \textbf{83(-16\%)} & 65(-33\%) & \textbf{96(-3\%)}  & 19(-80\%)          & \textbf{49(-24\%)} \\
0.6           & 0              & \textbf{100(0\%)} & \textbf{100(0\%)} & 85(-15\%)         & \textbf{100(0\%)}  & 76(-22\%) & \textbf{93(-6\%)}  & 8(-91\%)           & \textbf{59(-9\%)}  \\
0.6           & 0.03           & \textbf{100(0\%)} & 97(-2\%)          & 75(-25\%)         & \textbf{83(-16\%)} & 61(-37\%) & \textbf{86(-13\%)} & 9(-90\%)           & \textbf{39(-39\%)} \\
0.6           & 0.06           & \textbf{100(0\%)} & 91(-8\%)          & 76(-24\%)         & \textbf{80(-20\%)} & 60(-38\%) & \textbf{96(-3\%)}  & 10(-89\%)          & \textbf{46(-29\%)}\\
\bottomrule
\end{tabular}
\caption{Robustness analysis of our method and vanilla RL baseline. We evaluate the performance of agents on \textbf{training environments} under different noise levels, e.g., ``add 0.2 drop 0.03" means we randomly add 20\% additional edges while randomly dropping 3\% existing edges in graph. The numbers out of parentheses are normalized scores and the numbers in parentheses are relative performance change comparing to the performance without input noise. The bold numbers indicate which method performs better in each setting. }
\label{tab:textworld_robust}
\end{table}

As mentioned above, our method also aims at robustness. To evaluate the robustness of our models, we add different levels of noises to input knowledge graphs and regard the performance of the agents under noisy inputs. In this paper, we define the noise on a knowledge graph as adding additional edges to the graph or dropping existing edges in the graph. Formally, we add noise at $(k,p)$-level to input knowledge graph in the following way:

\begin{enumerate}
    \item Add edges: $\lceil k*|E| \rceil$ additional edges will be randomly generated and added to the graph, where $E$ is the edge set of the graph. For each edge $h_i, t_i, r_i$, head node $h_i$ and tail node $t_i$ are sampled from $V_{all}$ and relation $r_i$ is sampled from $R_{all}$, where $V_{all}$ means the set of all possible nodes and  $R_{all}$ is the set of all possible relations.
    \item Drop edges: $\lceil p*|E| \rceil$ edges will be dropped from the graph. Note that only original edges will be dropped, and the randomly added edges will not be dropped.
\end{enumerate}

Table \ref{tab:textworld_robust} shows the performance of vanilla RL and our method under noisy input graphs generated in the above mentioned way. Note that we test robustness in training environments instead of test environments, because the performance of both agents in test environments are not good enough and examining robustness of poor-performed agents does not make too much sense. 

We observed that in difficulty level 1, both agents are very robust under the noisy inputs. However, in difficulty level 2 \& 3 \& 4, the performance of the RL agents are hurts a lot by the input noises, especially in the difficulty level 4. In contrast, the agents obtained by our method are still performs pretty good and the performance drops significantly less than the RL agents.

{\color{black}
\subsection{Ablation Study}
\label{sec:ablation}

In this section, we study the contributions of different modules in our method. Our proposed method mainly consists of three modules: 1) an RL teacher policy, 2) a GNN-based action pruner, 3) a rule-based action selector. Therefore, we replace each module with other alternatives and see how the generalization performance is influenced. All the experiments below are performed on the TextWorld games with difficulty level 1. 

\subsubsection{RL teacher policy}

We first study how different kinds of teacher policies affects the performance. 
We consider the following three alternatives to our RL teacher policy: 1) an RL teacher policy with different backbone network: Graph Transformer Network (GTN) \cite{gtn}; 2) a manually written ground truth policy which can perfectly solves the tasks; 3) a random policy. Note that we choose ground truth policy and random policy as teachers because they represent two extreme cases: a very good teacher and a very bad teacher. 

\begin{table}[!htbp]
\centering
{\color{black}
\begin{tabular}{c|cc}
\toprule
Method Variant                     & Training & Test \\
\midrule
Ours (backbone is GCN)             & 100      & 100  \\
Use GTN as backbone of teacher     & 100      & 100  \\
Use ground truth policy as teacher & 100      & 100  \\
Use random policy as teacher       & 8.8      & 15.0 \\
\bottomrule
\end{tabular}
}
\caption{\color{black}Evaluation results on both training environments and test environment in TextWorld difficulty level 1. The numbers show the agent's normalized scores.}
\label{tab:ablation_teacher}
\end{table}

Table \ref{tab:ablation_teacher} shows the results of this ablation study. From this table, we can observe the following things:
\begin{itemize}
    \item It does not make much difference by switching the backbone of the RL teacher policy from GCN to GTN. This verifies our expectation: any graph-based network should be good as long as it has sufficient capacity for the training environment / data, because the job of the RL teacher is only to provide a demonstration dataset.
    \item Using ground truth policy also does not really affect the performance of our method. Since our RL teacher policy actually achieves almost perfect performance on the training environments, the demonstrations generated by our RL teacher and the ground truth policy are almost equally good.
    \item Using a random policy does not work at all. This also meets our expectation that a good demonstration dataset is very important for learning the action pruner and the rule-based action selector.
\end{itemize}

\subsubsection{GNN-based action pruner}

In this section, we mainly assess the influences of the network architectures used in the GNN-based action pruner. Specfically, we replace the backbone of our action pruner with Graph Transformer Network (GTN) \cite{gtn}. And we also study the case where the RL teacher also uses GTN as the backbone.

\begin{table}[!htbp]
\centering
{\color{black}
\begin{tabular}{c|cc}
\toprule
Method Variant                     & Training & Test \\
\midrule
Ours (GCN teacher, GCN action pruner)             & 100      & 100  \\
GCN teacher, GTN action pruner     & 100      & 100  \\
GTN teacher, GTN action pruner & 95      & 100  \\
\bottomrule
\end{tabular}
}
\caption{\color{black}Evaluation results on both training environments and test environment in TextWorld difficulty level 1. The numbers show the agent's normalized scores.}
\label{tab:ablation_pruner}
\end{table}

From the Table \ref{tab:ablation_pruner}, we can see that the architecture changes have minimal influence on the performance. This verifies our expectation again: any graph-based network should be good as long as it has sufficient capacity for the training environment / data. Though the “GTN teacher, GTN action pruner” variant does not get full scores on the training environment, the performance gap is very small.

\subsubsection{Rule-based action selector}

Finally, we study whether a rule-based action selector is crucial by replacing it with a network-based action selector. The resulting policy is somehow like a mixture of experts, where the gating function and the experts are all networks.

\begin{table}[!htbp]
\centering
{\color{black}
\begin{tabular}{c|cc}
\toprule
Method Variant                     & Training & Test \\
\midrule
Ours (rule-based action selector)             & 100      & 100  \\
GCN-based action selector     & 100      & 70  \\
\bottomrule
\end{tabular}
}
\caption{\color{black}Evaluation results on both training environments and test environment in TextWorld difficulty level 1. The numbers show the agent's normalized scores.}
\label{tab:ablation_selector}
\end{table}

From the Table \ref{tab:ablation_selector}, we can see that the network-based action selector does a good job in the training environments but performs poorly on the test environments. This is because the network-based action selector overfits the training data instead of really learning the underlying rules of decision making. In contrast, our rule-based action selector learns the real underlying rules that generalize well. Therefore, the rule-based action selector is a crucial component of our method.

}

%% file: sections/conclusion.tex
\section{Conclusion}
\label{sec:conclusion}
In this work, we propose a two-step hybrid policy for graph-based reinforcement learning. The two-step hybrid policy disentangles complicated black-box deep reinforcement learning policy into an action pruner and an action selector. The joint effort of two modules can not only generate human-friendly explanations of each decision, but also provide significantly better performance in terms of generalization and robustness. We conduct comprehensive experiments on the representative text-based environments~\cite{cote2018textworld} to demonstrate the effectiveness of our approach.